\documentclass[lettersize,journal]{IEEEtran}
\usepackage{graphicx}
\usepackage{amsmath}
\usepackage{algorithmicx,algorithm}
\usepackage[noend]{algpseudocode}
\usepackage{hyperref}
\hypersetup{hidelinks,
	colorlinks=true,
	allcolors=black,
	pdfstartview=Fit,
	breaklinks=true}
\usepackage{booktabs}
\usepackage{makecell}
\usepackage{multirow}
\usepackage{bm}

\newtheorem{myDef}{Definition}

\begin{document}

\title{Inducing Individual Students' Learning Strategies through Homomorphic POMDPs}

\author{Huifan Gao, Yifeng Zeng and Yinghui Pan


\thanks{Huifan Gao is with the Department of Automation, Xiamen University, Xiamen, China (e-mail:huifangao@stu.xmu.edu.cn).}

\thanks{Yifeng Zeng is with the Department of Computer \& Information Sciences, Northumbria University, UK (e-mail:yifeng.zeng@northumbria.ac.uk).}

\thanks{Yinghui Pan is with the College of Computer Science and Software Engineering, Shenzhen University, Shenzhen, China (e-mail: panyinghui@szu.edu.cn). 
}
}

\markboth{~Vol.~X,~No.~X,~month~yyyy}
{Shell \MakeLowercase{\textit{et al.}}: A Sample Article Using IEEEtran.cls for IEEE Journals}

\IEEEpubid{0000--0000~\copyright~2023 IEEE}

\maketitle

\begin{abstract}
Optimizing students' learning strategies is a crucial component in intelligent tutoring systems. Previous research has demonstrated the effectiveness of devising personalized learning strategies for students by modelling their learning processes through partially observable Markov decision process~(POMDP). However, the research holds the assumption that the student population adheres to a uniform cognitive pattern. While this assumption simplifies the POMDP modelling process, it evidently deviates from a real-world scenario, thus reducing the precision of inducing individual students' learning strategies. In this article, we propose the homomorphic POMDP~(H-POMDP) model to accommodate multiple cognitive patterns and present the parameter learning approach to automatically construct the H-POMDP model. Based on the H-POMDP model, we are able to represent different cognitive patterns from the data and induce more personalized learning strategies for individual students. We conduct experiments to show that, in comparison to the general POMDP approach, the H-POMDP model demonstrates better precision when modelling mixed data from multiple cognitive patterns. Moreover, the learning strategies derived from H-POMDPs exhibit better personalization in the performance evaluation.
\end{abstract}


\section{Introduction}
\label{sec:introduction}

\IEEEPARstart{A}{s} an application of artificial intelligence in the field of education, intelligent tutoring systems~(ITSs) have received increasing attention in recent years~\cite{wang2023examining}. The systems typically assist students in their learning pathway by providing learning materials~\cite{maciejewski1994student, vasandani1995knowledge, goh2006epilist}. Given the vast amount of learning resources available, selecting appropriate learning materials for students has been a focal point of the ITS research. This line of investigation primarily involves modelling the cognitive processes that occur when the students are engaged with the learning, thus deriving effective learning strategies.

Different from knowledge tracing~(KT) methods solely investigate the change of student knowledge states~\cite{pandey2020rkt, ghosh2020context, sun2021dynamic, su2021time}, the approach of partially observable Markov decision process~(POMDP)~\cite{spaan2012partially, fowler2022intelligent, liu2022false}, as a planable and interpretable temporal model, has shown promising results in improving the students' learning strategies in ITS~\cite{clement2016comparison, wang2018reinforcement, wang2019efficient}. Early research primarily explored the feasibility of utilizing POMDP for modelling the students' cognitive processes and inducing learning strategies through the students' learning activity data~\cite{rafferty2016faster}. Subsequent research has developed the enhancement to modelling methods and strategy induction techniques, and deal with complex knowledge domains~\cite{ramachandran2019personalized, nioche2021improving, gao2023improving}. However, the POMDP-based ITS approach still faces at least two main challenges in practical applications. On one hand, it often encounters difficulty in the model parameter learning particularly for the knowledge concepts with complex relationships. On the other hand, due to limited data on the learning activities, it is rather challenging to personalize individual learning strategies. Consequently, the POMDP-based  methods can only construct a general model from the data, which is not universally applicable to individual students.

The learning activity data generated during the learning process reflects the student's knowledge state and cognitive ability. However, the data is limited for a single student, and its scale is often insufficient to construct an accurate POMDP model. From a pedagogical perspective~\cite{judd2012educational}, the notions of {\it learning can be improved through student-to-student interaction} and {\it each student has their own suitable learning methods} reveal the principle that {\it cognitive abilities of different individuals exhibit both similarities and differences}. Thus, if it is possible to cluster learning data from students with similar abilities, allowing them to {\it exchange learning methods}, while differentiating the learning data of students with diverse abilities, enabling them to {\it maintain their individuality}, it becomes feasible to obtain more personalized learning strategies with a limited amount of data in the model learning. Similarly, there is a clustering problem involving time series with hidden states in the context of Hidden Markov Models~(HMM)~\cite{asadi2015creating}.

In this paper, drawing upon the research on the pedagogical methods~\cite{judd2012educational}, we aim to develop a personalized modelling approach for individual students. Specifically, we remove the assumption that all students shall adhere to a uniform cognitive pattern and propose a new modelling approach with the capability of accommodating multiple cognitive patterns. The new approach consists of multiple POMDPs, referred as homomorphic POMDPs~(H-POMDPs), that share the same state space, action space, and observation space, but have different parameter settings. Furthermore, we introduce a parameter learning method for the H-POMDP model, and by solving the learned models, we can obtain more personalized learning strategies.

In the real-time interaction with a student, using the H-POMDP model, we need to maintain beliefs not only about the student's knowledge states but also about the cognitive pattern to which the student belongs. Specifically, in each interaction, the belief about the student's cognitive pattern are updated first. For each cognitive pattern that the student may belong to, the belief about the student's knowledge state within that cognitive pattern need to be updated. Due to this particular hierarchical nature, the real-time strategy optimization~(i.e., determining which question can most effectively improve the student's knowledge) becomes more challenging. The main contributions of this article are summarized below.
\begin{itemize}
\item We propose a novel cognitive model called the H-POMDP model for modelling the cognitive processes of students' learning. In contrast to the general POMDP model that can only represent the average learning abilities of a student population, the new model can capture the patterns of how the students with different learning abilities update their knowledge states.
\item We develop an H-POMDP parameter learning method and an offline approach to solve the model. The learning approach differentiates different cognitive patterns each of which is used to learn one specific POMDP model.
\item We demonstrate the performance of the H-POMDP model in personalizing the students' learning strategies in two real-world knowledge concept learning domains.
\end{itemize}

The remaining sections of this paper are organized as follows. We start to present the POMDP background knowledge in Section~\ref{subsec:background} and provide a detailed exposition of the parameter learning and strategy induction methods for H-POMDP in Section~\ref{subsec:modelling} -Section~\ref{subsec:learning}. Section~\ref{sec:experiments} demonstrates the H-POMDP performance in the experiments. We review the related research of the POMDP-based cognitive models in Section~\ref{sec:related}. Section~\ref{sec:conclusion} summarizes this work and discusses further research.

\section{The Homomorphic POMDP Modelling Approach}
\label{sec:H-POMDP}

In this section, we start with background knowledge on the POMDP-based models in ITS and proceed to propose the new POMDP modelling as well as its parameter learning methods.

\subsection{Background Knowledge}
\label{subsec:background}

\begin{figure}[h]
    \centering
    \includegraphics[width=1\linewidth]{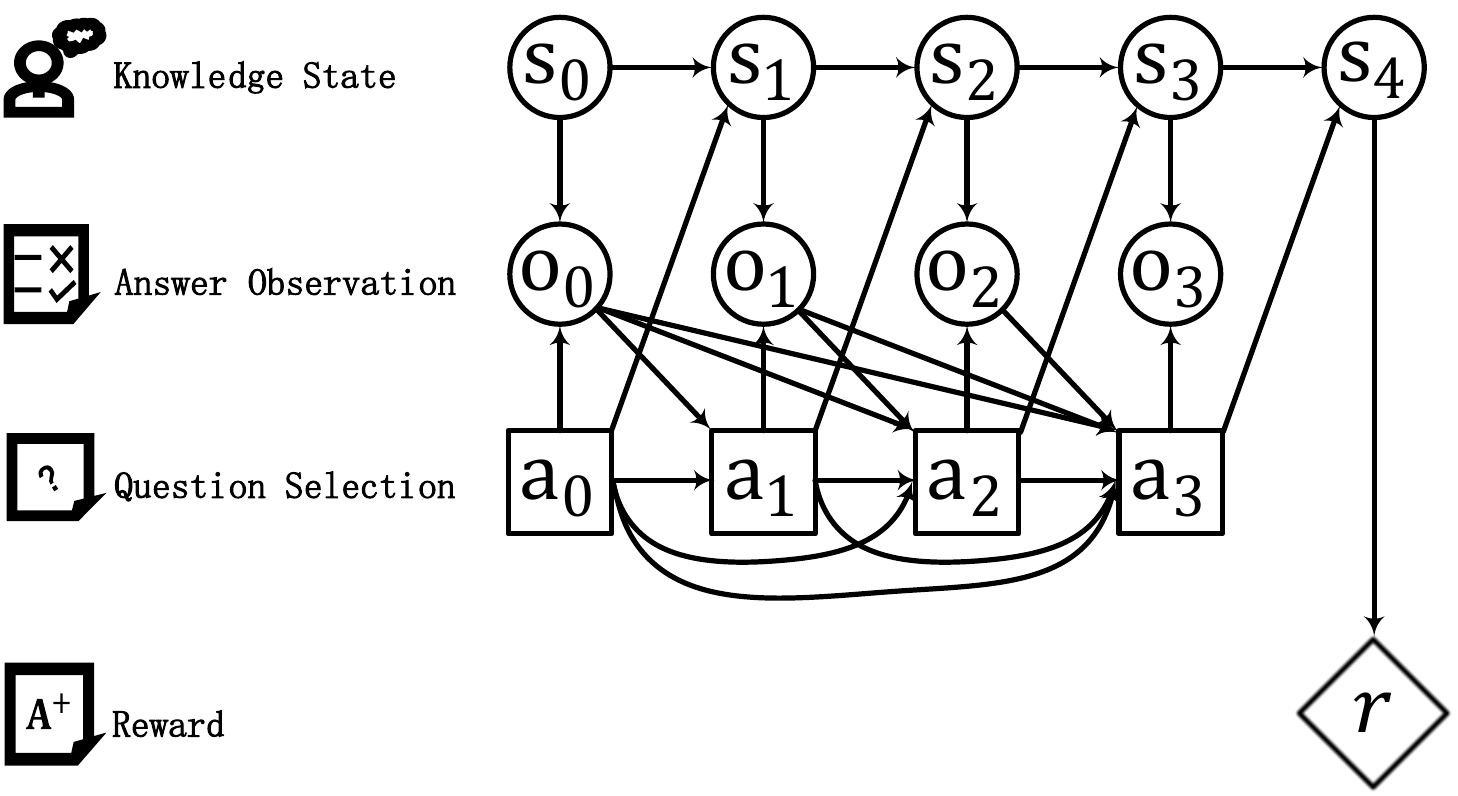}
    \caption{A POMDP-based cognitive model that plans the practice-based learning over four time steps.}
    \label{fig1}
\end{figure}

With the excellent interpretability, POMDP has been widely employed in modelling various scenarios in educational domains. We introduce a POMDP-based cognitive model, which is used to represent the student's practice-based learning process~\cite{gao2023improving}. In the practice-based learning process, answering questions serves as both a means to improve students' knowledge and a source of observations for a subsequent question selection. Fig.~\ref{fig1} illustrates a four-step POMDP model that captures the relationships among the student's knowledge state $s$, the question selection $a$ and the answer observation $o$.

Formally, given a learning domain that contains $N$ questions covering $K$ knowledge concepts of a specific structure, a POMDP-based cognitive model is a $7$-tuple $(S,A,T,\Omega,O,R,\gamma)$, where
\begin{itemize}
\item \textbf{$S$} is a set of knowledge states and each state $s\in{}S$ represents a specific mastery level across all $K$ knowledge concepts.
\item \textbf{$A$} is a set of actions and each action $a\in{}A$ is to select and answer a specific question in the knowledge learning domain.
\item \textbf{$T$} is the transition function $T(s'\mid{}s,a)$. It is the conditional probability of being in the new knowledge state $s'$ when the student selects and answers to a specific question in the the current knowledge state $s$. It characterizes how the student's knowledge mastery changes in the learning process.
\item \textbf{$\Omega$} is a set of observations, and each observation $o\in{}\Omega$ represents whether a student answers a specific question correctly or incorrectly.
\item \textbf{$O$} is the observation function $O(o|s,a)$. It is the conditional probability of seeing the observation $o$ when the student takes the action $a$~(selecting and answering to a specific question) in the knowledge state $s$.
\item \textbf{$R$} is the reward function. Its setting is related to the specific intelligent tutoring problem to be solved. For example, if the number of questions to be answered is preset, the reward can be only set for the final state $s_{final}$ and its value is determined by $s_{final}$.
\item \textbf{$\gamma\in\left[0,1\right]$} is a discount factor and limits the impact of actions in the future.
\end{itemize}

In order to facilitate the practical application of POMDP-based cognitive models in ITS, the existing research simplifies the model parameter learning and optimizes the learning strategies through educational knowledge~\cite{ebel1972essentials}. For example, by introducing a structure that represents the sequence of learning different knowledge concepts~\cite{millan2013using, wang2021training}, the POMDP state space can be effectively controlled. Using information entropy-based techniques~\cite{nunez1996information} to reduce the uncertainty in observing students' knowledge states contributes to improving the stability of learning strategies.

\subsection{H-POMDP Specification}
\label{subsec:modelling}

A single POMDP model is not sufficient to represent different cognitive patterns ascribed to students with different learning ability. Naturally we need a set of POMDP models that have different parameters while sharing identical structures including the state space, action space and their relations. We refer to the set of POMDPs as {\it homomorphic} POMDPs, namely H-POMDP. Formally, we can define the H-POMDP below.

\begin{myDef}
\label{def1}
Given a set of $k$ POMDP models, $\{m_1, m_2, \cdots, m_k\}$, that satisfy the following properties ($\Phi$, $\Psi$), we refer to the $k$ {\rm POMDP} models as an {\rm H-POMDP} model $M=\{m_1, m_2, \cdots, m_k\}$.
$$
\begin{aligned}
\Phi\leftarrow \{&S_{m_{1}}=S_{m_{2}}=\cdots=S_{m_{k}},\\&A_{m_{1}}=A_{m_{2}}=\cdots=A_{m_{k}},\\&\Omega_{m_{1}}=\Omega_{m_{2}}=\cdots=\Omega_{m_{k}}\}\\
\Psi \leftarrow \{&T_{m_{1}}\neq T_{m_{2}}\neq\cdots \neq T_{m_{k}},\\&O_{m_{1}}=O_{m_{2}}=\cdots=O_{m_{k}}=O,\\&R_{m_{1}}=R_{m_{2}}=\cdots=R_{m_{k}}=R\}
\end{aligned}
$$
\end{myDef}

In H-POMDP, all the $k$ models share the same state, action and observation spaces~($S$, $A$ and $\Omega$) and have the identical observation and reward functions~($O$ and $R$). They differ in the transition functions~($T$). Thus, H-POMDP can be used to describe the different dynamic patterns exhibited by different individuals or groups when performing the same task. We illustrate this concept through the scenario of practice-based learning. Suppose there are multiple students learning the same set of knowledge concepts through the practice-based learning. In this learning process, each student's knowledge state belongs to the same set of knowledge states~(state space $S$), and they can choose questions from the same question bank~(action space $A$) and only receive feedback about their knowledge state based on these questions~(observation space $\Omega$). However, for different students, the impact of practicing the knowledge concepts on their knowledge states would be different~(transition function $T$). In this scenario, the observation function $O$ is the same for all the students because whether they can answer the questions correctly is the criterion for assessing their mastery of the corresponding knowledge concepts, which needs to be standardized. To illustrate this point more specifically, let us consider a counterexample. Suppose two models \{$m_{1},m_{2}$\} have different observation functions $O_{1}$ and $O_{2}$. For a knowledge concept and a question related to it, in $m_{1}$~($m_{2}$), a student who has mastered this knowledge concept has a $80\%$~($70\%$) probability of answering the question correctly. Therefore, in these two models, the criteria for judging whether a student has mastered this knowledge concept are not the same - according to the criteria of $m_{1}$, compared to $m_{2}$, students need a higher rate of correct answers to be considered as having mastered this knowledge concept. This difference will lead to different actual meanings of the same knowledge state in different models, which conflicts with the property $S_{m_{1}}=S_{m_{2}}$. In other application scenarios, the observation function may vary - depending on the specific circumstances.

\subsection{H-POMDP Parameter Learning}
\label{subsec:parameter}

H-POMDP provides a POMDP-based cognitive framework where each POMDP model represents one type of cognitive pattern shared by a cohort of students. Given the set of learning activities data ${\rm H}=\left\{{\rm h}_{1},{\rm h}_{2},...,{\rm h}_{l}\right\}$, where ${\rm h}_{i}=\left\{({\rm a}_{i,1},{\rm o}_{i,1}),({\rm a}_{i,2},{\rm o}_{i,2}),...,({\rm a}_{i,{\rm T}_{i}-1},{\rm o}_{i,{\rm T}_{i}-1})\right\}$ is a sequence of actions and observations recorded in the data, we aim to learn both the transition and observation functions - a parameter learning process in H-POMDP.

Generally speaking, for a specific knowledge concept learning domain,  cognitive abilities of a student population can be categorized into $k$ cognitive patterns and each pattern models highly similar cognitive abilities. In other words, we can cluster the collective learning activity data into $k$ groups and each group encodes a specific cognitive pattern in one POMDP model within H-POMDP. Learning $k$ interacted POMDP models is not straightforward. The parameters of the $k$ cognitive patterns are not known prior to the learning, and need to be learned from the data. Learning these cognitive patterns requires the assignment of each data point into a specific cognitive pattern, which, in turn, demands to determine the specific parameters of each cognitive pattern. We note that the H-POMDP parameter learning involves a clustering problem where  the $k$ cognitive patterns are to be identified from a set of action and observation sequences $\rm H$.

We partition the set of learning activity data $\rm H$ into the observation sequences ${\rm {\bf O}}=\left\{{\rm O}_{1},{\rm O}_{2},...,{\rm O}_{l}\right\}$ and the action sequences ${\rm {\bf A}}=\left\{{\rm A}_{1},{\rm A}_{2},...,{\rm A}_{l}\right\}$. Assume the existence of $k$ patterns in the data, the objective is to learn the parameters $m=(D,T,O)$ for each pattern~($D$ is the distribution of the initial state). We integrate the parameters for each pattern into the H-POMDP parameters $M=\left\{m_{1},m_{2},...,m_{k}\right\}$. Let the corresponding state sequence be ${\rm {\bf S}}=\left\{{\rm S}_{1},{\rm S}_{2},...,{\rm S}_{l}\right\}$. Then, the generation probability for all observation sequences $\rm \bf O$ is given by
{\small
$$
\begin{aligned}
    P({\rm \bf O}\mid{}M,{\rm \bf A})=\mathop{\prod_{i=1}^{l}\sum_{j=1}^{k}\sum_{\rm S_{i}}}P({\rm O}_{i}\mid{\rm S}_{i},m_{j},{\rm A}_{i})P({\rm S}_{i}\mid{}m_{j},{\rm A}_{i})w_{i,j}
\end{aligned}
$$}
where $w_{i,j}$ is the probability~(referred to as a membership degree) of sequence ${\rm h}_i$ being generated by the pattern $m_j$. For simplicity, we consider the case of a single sequence, and later extend the results to multiple sequences.
$$
\begin{aligned}
    P({\rm O}\mid{}M,{\rm A})=\mathop{\sum_{j=1}^{k}\sum_{\rm S}}P({\rm O}\mid{\rm S},m_{j},{\rm A})P({\rm S}\mid{}m_{j},{\rm A})w_{i,j}
\end{aligned}
$$

We conduct the H-POMDP parameter learning  through the EM~(expectation and maximization) algorithm~\cite{dempster1977maximum} below.
\begin{itemize}
    \item Compute the log-likelihood of the complete data. We merge the observation sequence ${\rm O}=({\rm o}_{1},{\rm o}_{2},...,{\rm o}_{{\rm T}})$ and the state sequence ${\rm S}=({\rm s}_{1},{\rm s}_{2},...,{\rm s}_{{\rm T}})$ into the complete data $({\rm O}, {\rm S})=({\rm o}_{1},{\rm o}_{2},...,{\rm o}_{{\rm T}},{\rm s}_{1},{\rm s}_{2},...,{\rm s}_{{\rm T}})$. We then compute the log-likelihood function of the complete data as $logP({\rm O},{\rm S}\mid{}M,{\rm A})$.
    
    \item The E-step of the EM algorithm. We compute the Q-function $Q(M,w,\overline{M},\overline{w})$ below.
    {\scriptsize
    $$
    \begin{aligned}
    Q(M,w,\overline{M},\overline{w})=\mathop{\sum_{j=1}^{k}}\mathop{\sum_{\rm S}}\log{}[P({\rm O},{\rm S}\mid{}m_{j},{\rm A})w_{j}]P({\rm O},{\rm S}\mid{}\mathop{\overline{m_{j}}},{\rm A})\overline{w_{j}}
    \end{aligned}
    $$
    }
    where $\overline{M}=(\overline{m_{1}},\overline{m_{2}},...,\overline{m_{k}})$ and $\overline{w_{j}}$ are the current estimated values, and $M$ and $w_{j}$ are the parameters to be optimized.
    {\small
    $$
    \begin{aligned}
    P({\rm O},{\rm S}\mid{}m_{j},{\rm A})w_{j}=w_{j}D_{j}({\rm s_{1}})O({\rm o_{1}}\mid{}{\rm s_{1}},{\rm a_{1}})T_{j}({\rm s_{2}}\mid{}{\rm s_{1}},{\rm a_{1}})\\O({\rm o_{2}}\mid{}{\rm s_{2}},{\rm a_{2}})T_{j}({\rm s_{3}}\mid{}{\rm s_{2}},{\rm a_{2}}){\cdots{}}\\O({\rm o_{T-1}}\mid{}{\rm s_{T-1}},{\rm a_{T-1}})T_{j}({\rm s_{T}}\mid{}{\rm s_{T-1}},{\rm a_{T-1}})
    \end{aligned}
    $$
    }
    Hence, the function $Q(M,w,\overline{M},\overline{w})$ can be expressed as
    {\small
    \begin{flalign}
    \begin{split}
    Q(M,w,\overline{M},\overline{w})=\mathop{\sum_{j=1}^{k}\sum_{\rm S}}\log{}w_{j}P({\rm O},{\rm S}\mid{}\overline{m_{j}},{\rm A})\overline{w_{j}}+\\\mathop{\sum_{j=1}^{k}\sum_{\rm S}}\log{}D_{j}({\rm s_{1}})P({\rm O},{\rm S}\mid{}\overline{m_{j}},{\rm A})\overline{w_{j}}+\\\mathop{\sum_{j=1}^{k}\sum_{\rm S}[\sum_{t=1}^{\rm T-1}}\log{}T_{j}({{\rm s}_{t+1}}\mid{}{{\rm s}_{t},{\rm a}_{t}})]P({\rm O},{\rm S}\mid{}\overline{m_{j}},{\rm A})\overline{w_{j}}+\\\mathop{\sum_{j=1}^{k}\sum_{\rm S}[\sum_{t=1}^{\rm T-1}}\log{}O({{\rm o}_{t}}\mid{}{{\rm s}_{t}},{{\rm a}_{t}})]P({\rm O},{\rm S}\mid{}\overline{m_{j}},{\rm A})\overline{w_{j}}
    \end{split}
    \label{eq1}
    \end{flalign}
    }
    The summations are taken over the total sequence length ${\rm T}$.

    \item The M-step of the EM algorithm. We maximize the Q-function $Q(M,w,\overline{M},\overline{w})$ to obtain the model parameters $D$, $T$, $O$, and $w$. Since the parameters to be optimized appear separately in each of the four terms in Eq.~\ref{eq1}, we can perform the maximization for each term in a separate way.

    \noindent{}($a$) The first term can be expressed as
    {\footnotesize
    $$
    \begin{aligned}
    \mathop{\sum_{j=1}^{k}\sum_{\rm S}}\log{}w_{j}P({\rm O},{\rm S}\mid{}\overline{m_{j}},{\rm A})\overline{w_{j}}=\mathop{\sum_{j=1}^{k}}\log{}w_{j}P({\rm O}\mid{}\overline{m_{j}},{\rm A})\overline{w_{j}}
    \end{aligned}
    $$
    }
    Note that $w_{j}$ satisfies the constraint $\sum_{j=1}^{k}w_{j}=1$. Using the {\it Lagrange} multipliers, we write the {\it Lagrange} function as
    $$
    \begin{aligned}
    \mathop{\sum_{j=1}^{k}}\log{}w_{j}P({\rm O}\mid{}\overline{m_{j}},{\rm A})\overline{w_{j}}+\gamma(\mathop{\sum_{j=1}^{k}}w_{j}-1)
    \end{aligned}
    $$
    We take the partial derivatives and set the results to $0$.
    $$
    \begin{aligned}
    \frac{\partial}{\partial w_{j}}[\mathop{\sum_{j=1}^{k}}\log{}w_{j}P({\rm O}\mid{}\overline{m_{j}},{\rm A})\overline{w_{j}}+\gamma(\mathop{\sum_{j=1}^{k}}w_{j}-1)]=0
    \end{aligned}
    $$
    \begin{flalign}
    P({\rm O}\mid{}\overline{m_{j}},{\rm A})\overline{w_{j}}+\gamma{}w_{j}=0
    \label{eq2}
    \end{flalign}
    We can sum over $j$ to obtain $\gamma$ in Eq.~\ref{eq2}.
    $$
    \begin{aligned}
    \gamma=-\mathop{\sum_{j=1}^{k}}P({\rm O}\mid{}\overline{m_{j}},{\rm A})\overline{w_{j}}
    \end{aligned}
    $$
    By substituting it into Eq.~\ref{eq2}, we obtain the $w_j$ value.
    \begin{flalign}
    w_{j}=\frac{P({\rm O}\mid{}\overline{m_{j}},{\rm A})\overline{w_{j}}}{\sum_{j'=1}^{k}P({\rm O}\mid{}\overline{m_{j'}},{\rm A})\overline{w_{j'}}}
    \label{eq3}
    \end{flalign}
    \noindent{}($b$) The second term can be expressed as
    $$
    \begin{aligned}
    \mathop{\sum_{j=1}^{k}\sum_{\rm S}}\log{}D_{j}({\rm s_{1}})P({\rm O},{\rm S}\mid{}\overline{m_{j}},{\rm A})\overline{w_{j}}=\\\mathop{\sum_{j=1}^{k}\sum_{{\rm s}=s_{0}}^{s_{\mid{}S\mid{}}}}\log{}D_{j}({\rm s_{1}=s})P({\rm O},{\rm s_{1}=s}\mid{}\overline{m_{j}},{\rm A})\overline{w_{j}}
    \end{aligned}
    $$
    Similar to the first term, we apply the {\it Lagrange} multiplier method with the constraint $\sum_{{\rm s}={s_{0}}}^{s_{\mid{}S\mid{}}}D_{j}({\rm s_{1}=s})=1$.
    \begin{flalign}
    D_{j}({\rm s_{1}=s})=\frac{P({\rm O},{\rm s_{1}=s}\mid{}\overline{m_{j}},{\rm A})\overline{w_{j}}}{P({\rm O}\mid{}\overline{m_{j}},{\rm A})\overline{w_{j}}}
    \label{eq4}
    \end{flalign}
    \noindent{}($c$) The third term can be expressed as
    {\scriptsize
    $$
    \begin{aligned}
    \mathop{\sum_{j=1}^{k}\sum_{\rm S}[\sum_{t=1}^{\rm T-1}}\log{}T_{j}({{\rm s}_{t+1}}\mid{}{{\rm s}_{t},{\rm a}_{t}})]P({\rm O},{\rm S}\mid{}\overline{m_{j}},{\rm A})\overline{w_{j}}=\\\mathop{\sum_{j=1}^{k}\sum_{{\rm s}=s_{0}}^{s_{\mid{}S\mid{}}}\sum_{{\rm s'}=s_{0}}^{s_{\mid{}S\mid{}}}\sum_{t=1}^{\rm T-1}}\log{}T_{j}({\rm s'}\mid{}{\rm s,a})P({\rm O},{{\rm s}_{t}=s},{{\rm s}_{t+1}=s'}\mid{}\overline{m_{j}},{\rm A})\overline{w_{j}}
    \end{aligned}
    $$
    }
    We apply the {\it Lagrange} multiplier method with the constraint $\sum_{{\rm s'}=s_{0}}^{s_{\mid{}S\mid{}}}T_{j}({\rm s'\mid{}s,a})=1$~(only when ${\rm a}_{t}={\rm a}$, the partial derivative of $T_{j}({{\rm s}_{t+1}}\mid{}{{\rm s}_{t},{\rm a}_{t}})$ with respect to $T_{j}({\rm s'}\mid{}{\rm s,a})$ is not $0$, denoted as $I({\rm a}_{t}={\rm a})$).
    {\footnotesize
    \begin{flalign}
    T_{j}({\rm s'}\mid{}{\rm s,a})=\frac{\sum_{t=1}^{\rm T-1}P({{\rm O}, {\rm s}_{t}=s, {\rm s}_{t+1}=s'}\mid{}\overline{m_{j}},{\rm A})\overline{w_{j}}I({\rm a}_{t}={\rm a})}{\sum_{t=1}^{\rm T-1}P({{\rm O}, {\rm s}_{t}=s}\mid{}\overline{m_{j}},{\rm A})\overline{w_{j}}I({\rm a}_{t}={\rm a})}
    \label{eq5}
    \end{flalign}
    }
    \noindent{}($d$) The fourth term can be expressed as
    {\small
    \begin{flalign}
    \begin{split}
    \mathop{\sum_{j=1}^{k}\sum_{\rm S}[\sum_{t=1}^{\rm T-1}}\log{}O({\rm o}_{t}\mid{}{\rm s}_{t},{\rm a}_{t})]P({\rm O},{\rm S}\mid{}\overline{m_{j}},{\rm A})\overline{w_{j}}=\\\mathop{\sum_{j=1}^{k}\sum_{{\rm s}=s_{0}}^{s_{\mid{}S\mid{}}}\sum_{t=1}^{\rm T-1}}\log{}O({\rm o}_{t}\mid{}{\rm s},{\rm a})P({\rm O},{\rm s}_{t}=s\mid{}\overline{m_{j}},{\rm A})\overline{w_{j}}
    \end{split}
    \label{eq6}
    \end{flalign}
    }
    Similarly, we use the {\it Lagrange} multiplier method with the constraint $\sum_{{\rm o}=o_{0}}^{o_{\mid{}O\mid{}}}O({\rm o\mid{}s,a})=1$~(only when ${\rm a}_{t}={\rm a}$ and ${\rm o}_{t}={\rm o}$, the partial derivative of $O({\rm o}_{t}\mid{}s,a)$ with respect to $O({\rm o\mid{}s,a})$ is not $0$, denoted as $I({\rm a}_{t}={\rm a})$ and $I({\rm o}_{t}={\rm o})$).
    {\scriptsize
    \begin{flalign}
    O({\rm o\mid{}s,a})=\frac{\sum_{j=1}^{k}\sum_{t=1}^{\rm T-1}P({\rm O,s}_{t}=s\mid{}\overline{m_{j}},{\rm A})\overline{w_{j}}I({\rm a}_{t}={\rm a})I({\rm o}_{t}={\rm o})}{\sum_{j=1}^{k}\sum_{t=1}^{\rm T-1}P({\rm O,s}_{t}=s\mid{}\overline{m_{j}},{\rm A})\overline{w_{j}}I({\rm a}_{t}={\rm a})}
    \label{eq7}
    \end{flalign}
    }

    \item By extending the results of Eq.~\ref{eq3},~\ref{eq4},~\ref{eq5} and~\ref{eq7} to the case of multiple sequences, we update the H-POMDP parameter below.
    \begin{flalign}
    w_{i,j}=\frac{P({\rm O}_{i}\mid{}\overline{\lambda_{j}},{\rm A}_{i})}{\sum_{j'=1}^{k}P({\rm O}_{i}\mid{}\overline{\lambda_{j'}},{\rm A}_{i})}
    \label{eq8}
    \end{flalign}
    \begin{flalign}
    D_{j}({\rm s_{1}=s})=\frac{1}{l}\sum_{i=1}^{l}\frac{P({\rm O}_{i},{\rm s}_{i,1}={\rm s}\mid{}\overline{\lambda_{j}},{\rm A}_{i})}{P({\rm O}_{i}\mid{}\overline{\lambda_{j}},{\rm A}_{i})}
    \label{eq9}
    \end{flalign}
    {\fontsize{7.75}{12}
    \begin{flalign}
    \begin{split}
    &T_{j}({\rm s'}\mid{}{\rm s,a})\\=&\frac{\sum_{i=1}^{l}\sum_{t=1}^{{\rm T}_{i}-1}P({\rm O}_{i}, {\rm s}_{i,t}={\rm s}, {\rm s}_{i,t+1}={\rm s'}\mid{}\overline{\lambda_{j}},{\rm A}_{i})\overline{w_{i,j}}I({\rm a}_{i,t}={\rm a})}{\sum_{i=1}^{l}\sum_{t=1}^{{\rm T}_{i}-1}P({\rm O}_{i}, {\rm s}_{i,t}={\rm s}\mid{}\overline{\lambda_{j}},{\rm A}_{i})\overline{w_{i,j}}I({\rm a}_{i,t}={\rm a})}
    \end{split}
    \label{eq10}
    \end{flalign}
    }
    {\fontsize{6}{12}
    \begin{flalign}
    \begin{split}
    &O({\rm o\mid{}s,a})\\=&\frac{\sum_{i=1}^{l}\sum_{j=1}^{k}\sum_{t=1}^{{\rm T}_{i}-1}P({\rm O}_{i},{\rm s}_{i,t}={\rm s}\mid{}\overline{\lambda_{j}},{\rm A}_{i})\overline{w_{i,j}}I({\rm a}_{i,t}={\rm a})I({\rm o}_{i,t}={\rm o})}{\sum_{i=1}^{\rm l}\sum_{j=1}^{k}\sum_{t=1}^{{\rm T}_{i}-1}P({\rm O}_{i},{\rm s}_{i,t}={\rm s}\mid{}\overline{\lambda_{j}},{\rm A}_{i})\overline{w_{i,j}}I({\rm a}_{i,t}={\rm a})}
    \end{split}
    \label{eq11}
    \end{flalign}
    }
\end{itemize}

\begin{figure*}[h]
    \centering
    \includegraphics[width=1.\linewidth]{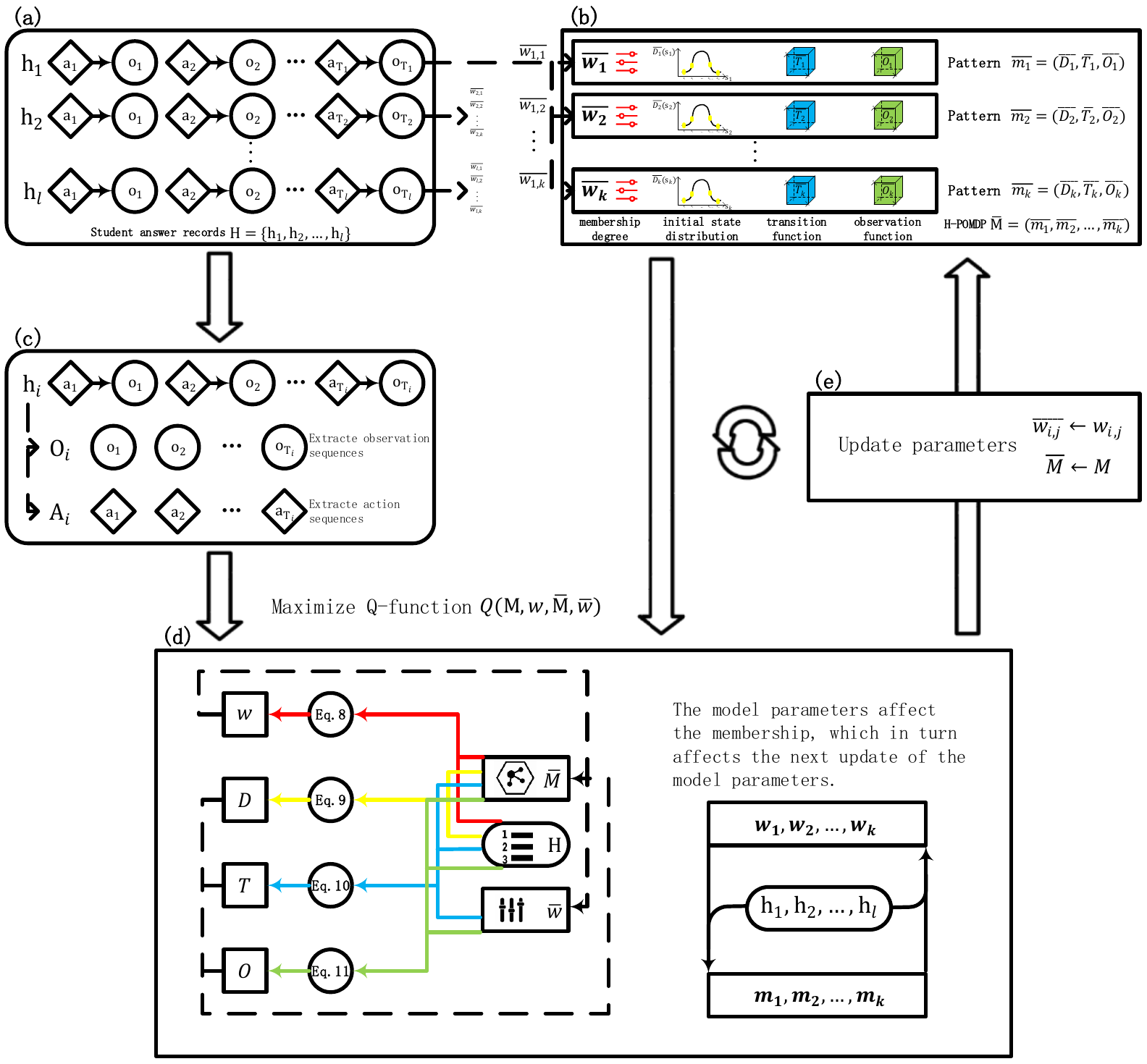}
    \caption{The H-POMDP parameter learning contains the procedures~ ($\rm a$-$\rm e$) where the sequences are clustered and the parameter values are updated simultaneously.}
    \label{fig2}
\end{figure*}

As shown in Fig.~\ref{fig2}, we start by initializing the membership degree of each sequence to each cognitive pattern that is represented by one POMDP model~($\rm a$) and the parameters of $k$ POMDPs~($\rm b$). In the update formulations, the observation sequence and the action sequence are separated. Thus, for each sequence, we extract its observation sequence and action sequence separately to facilitate the subsequent parameter updates~($\rm c$). During the parameter update, the membership degree is updated based on the current estimated values of the model parameters and the sequences, similar to reclassifying each data sequence after confirming new clustering centers; and the new membership degree affects the weight of each sequence in the next parameter update, which is similar to the re-calculation of clustering centers~($\rm d$). We iterate this update process until the parameter values converge~(e.g. the difference between the parameter values of the last two updates does not exceed a certain threshold)~($\rm e$). Through this method, we can simultaneously learn the model parameters and cluster the sequences.

We summarize the above procedures in Alg.~\ref{algo1}. Given available learning records, we generate the state space $S$, action space $A$, and observation space $\Omega$ based on the specific problem to be solved~(line 1), and randomly initialize the parameters $M$ and $w$~(line 2). Then, we use the EM algorithm to update the parameters~(line 3-16) until the termination condition is met~(e.g. the error between the parameters of the last two updates does not exceed a certain threshold).\begin{algorithm}[h]
    \caption{Parameter Learning of H-POMDP}
    \hspace*{0.02in} {\bf Input:}The observation sequences ${\rm \bf O}$, the action sequences ${\rm \bf A}$, the number of patterns $k$, the termination condition \\
    \hspace*{0.02in} {\bf Output:}The membership $w$, the initial state distribution $D$, the transition function $T$, and the observation function $O$.
    \begin{algorithmic}[1]
        \State{Generate $S$, $A$, and $\Omega$.} 
        \State{Randomly initialize $w$ and $M$.}
        \While{the termination condition is not met}
            \For{$j\in{}\left\{1,2,...,k\right\}$}
                \For{$i\in{}\left\{1,2,...,{\rm l}\right\}$}
                    \State{$\overline{w_{i,j}}\overline{}\leftarrow{}w_{i,j}$}
                \EndFor
                \State{$\overline{m_{j}}\leftarrow{}m_{j}$}
            \EndFor
            \For{$j\in{}\left\{1,2,...,k\right\}$}
                \For{$i\in{}\left\{1,2,...,{\rm l}\right\}$}
                    \State{Calculate $w_{i,j}$ according to Eq.~\ref{eq8}}
                \EndFor
                \For{${\rm s}\in{}S$}
                    \State{Calculate $D_{j}({\rm s_{1}=s})$ according to Eq.~\ref{eq9}}
                \EndFor
                \For{${\rm s,s'}\in{}S,{\rm a}\in{}A$}
                    \State{Calculate $T_{j}({\rm s'}\mid{}{\rm s,a})$ according to Eq.~\ref{eq10}}
                \EndFor
            \EndFor
            \For{${\rm s}\in{}S,{\rm o}\in{}\Omega,{\rm a}\in{}A$}
                \State{Calculate $O({\rm o\mid{}s,a})$ according to Eq.~\ref{eq11}}
            \EndFor
        \EndWhile
       \State{{\bf return} $w,D,T,O$}
    \end{algorithmic}
    \label{algo1}
\end{algorithm}

We shall note that the H-POMDP parameter learning is not simply equivalent to learning each POMDP model separately from one portion of learning activity records. We conduct the learning and clustering processes simultaneously and update the H-POMDP parameters using all the information, e.g. the model memberships, from the $k$ models.

\subsection{Simplification in the Parameter Learning}
\label{subsec:simplification}

As H-POMDP is a specific application for students' knowledge concept learning, there exist some common educational principles~\cite{judd2012educational}. We can use these principles as constraints on learning the model parameters so as to prevent unrealistic patterns. This allows the model to induce learning strategies that are consistent with educational principles. Thus, we impose a set of reasonable constraints based on the practical meanings of the parameters. The constraint will also simplify the parameter learning of the transition and observation functions in H-POMDP.

{\it Constraint 1. Given a knowledge concept, the students who have mastered the concept have a larger probability of answering questions related to the concept correctly compared to the students who have not mastered the concept.}

This constraint is evident and can be expressed as
$$
\begin{aligned}
    O(1_{a}|s,a)>O(1_{a}|s',a)
    \label{eq23}
\end{aligned}
$$
where the knowledge concept is~(not) mastered in state $s$~($s'$). For a knowledge concept, the students have a probability of answering questions correctly through guessing even when they have not mastered it. Conversely, when they have mastered it, there is still a chance of making mistakes in their answers. Suppose there is insufficient data, the result of parameter learning may be that the student has a higher probability of guessing the correct answer than correctly answering the question given their mastery of the corresponding knowledge concept. To avoid learning such illogical parameters, we introduce this constraint to the observation function. Adding this constraint is a straightforward extension in the relevant derivation of Eq.~\ref{eq6}.

{\it Constraint 2. Given a student's knowledge state, the impact of answering questions related to the same knowledge concept is consistent.}

In H-POMDP, the proficiency of students in understanding knowledge concepts is roughly modelled as {\it mastered} or {\it non-mastered}. It does not capture the finer proficiency levels which, of course, is difficult to be modelled precisely. The impact of answering questions  primarily occurs when the students rectify their misconceptions and reinforce correct understanding after checking the answers. Thus, we have the constraint
$$
\begin{aligned}
    T(s'\mid{}s,a)=T(s'\mid{}s,a')
    \label{eq24}
\end{aligned}
$$
where the questions $a$ and $a'$ are related to the same knowledge concept. Furthermore, the difficulty level of questions may not necessarily stem from the inherent complexity of the knowledge concept, but rather from intricate problem-solving techniques required to answer the questions. In such cases, the probability of correctly answering the questions might be low, but it does not imply that the questions do not contribute to the learning of the associated knowledge concept.

{\it Constraint 3. For a knowledge concept, a student who has not mastered it can only acquire mastery by answering questions related to the concept, and a student who has already mastered the concept will not immediately forget it after answering questions related to it.}

This constraint shows that the change in students' knowledge state is closely related to their learning activities~\cite{solso2005cognitive}. Based on this constraint, we can set the relevant transition probabilities to $0$, and the transition probabilities with a value of $0$ are not actually involved in the iterative parameter learning. We represent this constraint as
$$
\begin{aligned}
    T(s'\mid{}s,a)=0\;\;\;T(s\mid{}s',a')=0
\end{aligned}
$$
where $s$~($s'$) is the knowledge state related to the knowledge concept that has~(not) been mastered and $a$~($a'$) is the question selection that does~(not) involve the knowledge concept. In this scenario, if a student alternately engage with several different knowledge concepts in the learning, the parameter learning may fail to distinguish the reasons for changing his/her knowledge states in the situations with limited data. Hence, we need to introduce this constraint to prevent unreasonable {\it knowledge concept forgetting} and {\it knowledge concept acquisition}.

The three constraints are commonly identified in the students' learning pathway and reduce the computational complexity in the H-POMDP parameter learning.

\subsection{Learning Strategy Induction}
\label{subsec:learning}

Despite its special structure, H-POMDP can still be solved using dynamic programming~(DP) method~\cite{bellman1966dynamic} and the optimal learning strategy is to be induced from the H-POMDP solution. Unlike an ordinary POMDP, the H-POMDP model requires the maintenance of both the pattern belief $bm$ and the state belief $bs$.

We can not definitively know a student's cognitive pattern, but can estimate his/her potential membership in various cognitive patterns in a probabilistic manner. If the student has no prior answer records, we can only represent his/her initial pattern belief and state belief using the average levels of the student population based on the historical data.
$$
\begin{aligned}
    bm_{1}(m) = \frac{1}{l}\sum_{i=1}^{l}w_{i,m}\;\;\;bs_{1}({\rm s_{1}}\mid{}m) = D_{m}({\rm s_{1}})
\end{aligned}
$$
Subsequently, we update the pattern and the state beliefs as follows.
\begin{flalign}
    \resizebox{.885\hsize}{!}{$bm_{t}(m)=\frac{\sum_{s_{t-1}\in{}S}bm_{t-1}(m)bs_{t-1}(s_{t-1}\mid{}m)O(o_{t-1}\mid{}s_{t-1},a_{t-1})}{\sum_{m'\in{}M}\sum_{s_{t-1}\in{}S}bm_{t-1}(m')bs_{t-1}(s_{t-1}\mid{}m')O(o_{t-1}\mid{}s_{t-1},a_{t-1})}$}
    \label{eq12}
\end{flalign}
\begin{flalign}
    \resizebox{.885\hsize}{!}{$bs_{t}(s_{t}\mid{}m)=\frac{\sum_{s_{t-1}\in{}S}bs_{t-1}(s_{t-1}\mid{}m)O(o_{t-1}|s_{t-1},a_{t-1})T_{m}(s_{t}|s_{t-1},a_{t-1})}{\sum_{s'_{t}\in{}S}\sum_{s_{t-1}\in{}S}bs_{t-1}(s_{t-1}\mid{}m)O(o_{t-1}|s_{t-1},a_{t-1})T_{m}(s'_{t}|s_{t-1},a_{t-1})}$}
    \label{eq13}
\end{flalign}

Upon the current pattern and state belief $bm_{t}(m)$ and $bs_{t}(s_{t}\mid{}m)$, the optimal strategy $\pi^{\ast}(bm_{t}, bs_{t})$ is to choose the action with the highest action value
$$
\begin{aligned}
    \pi^{\ast}(bm_{t},bs_{t})=argmax_{a_{t}\in{}A}Q(bm_{t},bs_{t},a_{t})
\end{aligned}
$$
where the action value is calculated as
$$
\begin{aligned}
    \resizebox{1.\hsize}{!}{$Q(bm_{t},bs_{t},a_{t})=R(bm_{t},bs_{t},a_{t})+\gamma{}{\sum}_{o_{t}\in{}\Omega}O(o_{t}\mid{}bm_{t},bs_{t},a_{t})V_{t+1}(bm_{t+1},bs_{t+1})$}
\end{aligned}
$$
where the next pattern belief $bm_{t+1}$ and the next state belief $bs_{t+1}$ are updated according to Eq.~\ref{eq12} and Eq.~\ref{eq13} respectively.

The belief value $V_{t}(bm_{t},bs_{t})$ is the value of the action that maximizes the new belief value therefore leading to the recursive computation in the DP
$$
\begin{aligned}
    V_{t}(bm_{t},bs_{t})=max_{a_{t}\in{}A}Q(bm_{t},bs_{t},a_{t})
\end{aligned}
$$

\section{Experimental Results}
\label{sec:experiments}

We implement the H-POMDP parameter learning technique in Alg.~\ref{algo1} and the learning strategy induction method in Section~\ref{subsec:learning}. In addition, we implement an ordinary POMDP method and learn the single POMDP from the data through the Baum-Welch algorithm~\cite{baum1970maximization}. We conduct a series of experiments in two tasks, namely the student performance prediction and learning strategy induction, to demonstrate the H-POMDP performance over the baseline POMDP model. All the tests are conducted through Python 3.8 on an Ubuntu server with a Core i9-1090K 3.7GHz, a GeForce RTX 3090 and a 128 GB memory.

\subsection{Datasets on the Learning Activity}

\begin{figure}[h]
    \centering
    \includegraphics[width=1.\linewidth]{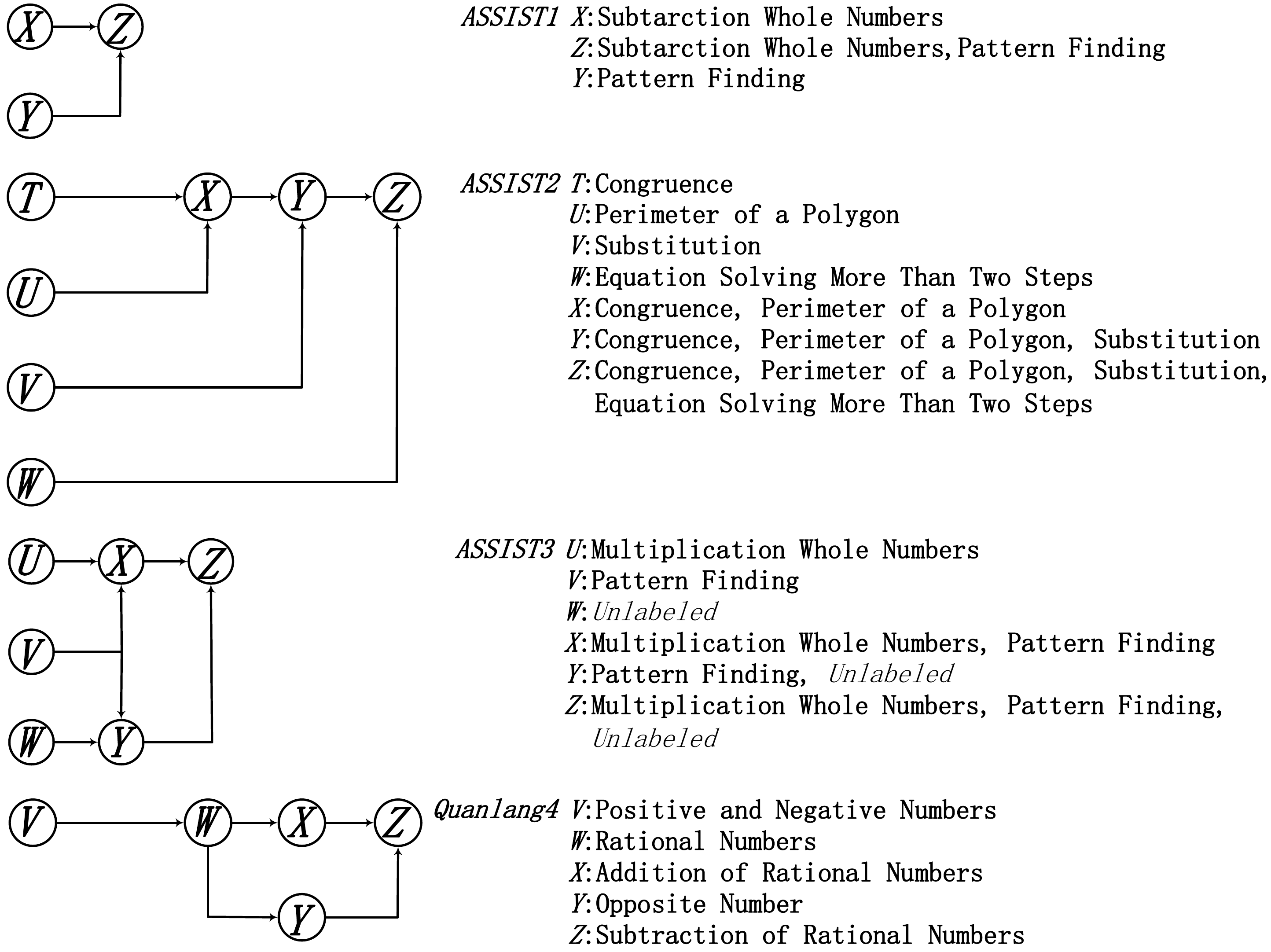}
    \caption{Knowledge concept structures of the four sub-datasets. The records of {\it ASSIST1}, {\it ASSIST2} and {\it ASSIST3} are from {\it ASSIST}, and the records of {\it Quanlang1} are from {\it QuanLang}.}
    \label{fig3}
\end{figure}

We extract four sub-datasets~({\it ASSIST1}, {\it ASSIST2}, {\it ASSIST3}, {\it Quanlang1})  from two large datasets, including the publicly available dataset ASSIST~(Non Skill-Builder data 2009-10)~\cite{feng2009addressing} and the private dataset {\it Quanlang}, based on relationships among knowledge concepts. {\it ASSIST} is the publicly available data collected by the ASSISTments online tutoring system during the 2009-2010 academic year. {\it Quanlang} is the data from the middle schools that have partnerships with the {\it Quanlang} education company~\footnote{\href{https://www.quanlangedu.com}{https://www.quanlangedu.com}}. Each dataset includes students' answer records and relevant question information. We provide a brief summary of these datasets and sub-datasets in Table~\ref{tbl2}.\begin{table}[h]
    \renewcommand{\arraystretch}{1}
    \scriptsize
    \setlength\tabcolsep{5pt}
    \caption{The statistical features of the datasets and sub-datasets in the experiments}
    \begin{tabular}{lrrrr}
        \toprule
        \textbf{dataset}/sub-dataset & Students & Knowledge concepts & Questions & Answer logs \\
        \midrule
        \textbf{\textit{ASSIST}} & 8,096 & 200 & 6,907 & 603,128\\
        \textit{ASSIST1} & 2,865 & 3 & 202 & 25,963\\
        \textit{ASSIST2} & 3,512 & 7 & 157 & 21,171\\
        \textit{ASSIST3} & 3,283 & 6 & 208 & 25,642\\
        \\
		\textbf{\textit{Quanlang}} & 11,203 & 14 & 3,544 & 243,718\\
        \textit{Quanlang1} & 9,462 & 5 & 1,288 & 96,827\\
        \bottomrule
    \end{tabular}
    \label{tbl2}
\end{table}
The knowledge concept structures corresponding to each sub-dataset, as depicted in Fig.~\ref{fig3}, are either obtained from the information provided on the {\it ASSIST} website or from domain experts at {\it Quanlang}.

Details about {\it ASSIST} can be found in reference~\cite{feng2009addressing}. Here, we provide a brief introduction to {\it Quanlang}. Table~\ref{tbl3} displays the some samples of the answer records. In our experiments, we primarily utilized the attributes {\it seq\_id}, {\it account\_no}, {\it exam\_id}, {\it knowledge\_concept\_id} and {\it is\_right}. Table~\ref{tbl4} explains these attributes.\begin{table}[h]
    \renewcommand{\arraystretch}{1}
    \setlength\tabcolsep{1.75pt}
    \tiny
    \caption{Record examples of {\it Quanlang}}
    \begin{tabular}{rrrrrrrrrrr}
        \toprule
        seq\textbf{\_}id & task\textbf{\_}id & account\textbf{\_}no & exam\textbf{\_}id & knowledge\textbf{\_}concept\textbf{\_}id & auto\textbf{\_}scoring & result & is\textbf{\_}right & is\textbf{\_}mark & score & class\textbf{\_}id \\
        \midrule
        1845890 & 40243 & 10446 & 19475 & 10762 & 0 & ``70m" & I01 & I01 & 5 & 322 \\
        1864626 & 41737 & 11398 & 19481 & 70762 & 0 & ``4" & I02 & I01 & 2 & 322 \\
        12016 & 1909 & 12068 & 19432 & 10762 & 1 & ``D" & I03 & I01 & 0 & 60 \\
        \bottomrule
    \end{tabular}
    \label{tbl3}
\end{table}\begin{table}[h]
    \renewcommand{\arraystretch}{1}
    \setlength\tabcolsep{5pt}
    \scriptsize
    \caption{Elaboration of some data attributes in $Quanlang$}
    \begin{tabular}{ll}
        \toprule
        Field & Annotation \\
        \midrule
        seq\textbf{\_}id & It uniquely identifies the record in chronological order.\\
        \\
        account\textbf{\_}no & student account \\
        \\
        exam\textbf{\_}id & It uniquely identifies a question.\\
        \\
        knowledge\textbf{\_}concept\textbf{\_}id & It uniquely identifies a knowledge concept.\\
        \\
        is\textbf{\_}right & \makecell[l]{It indicates whether the student's answer is correct. I01\\ denotes complete correctness, I02 denotes partial correct-\\ness, and I03 denotes incorrect.}\\
        \\
        \bottomrule
    \end{tabular}
    \label{tbl4}
\end{table}

\subsection{Experiment 1: Evaluations on Students' Performance Prediction}

We first validate the superior performance of the homomorphic POMDPs cognitive model compared to the POMDP-based cognitive model through the students' performance prediction task.

For any sequence in the dataset, the student performance prediction task is to predict the student's next response based on the known conditions, which include the student's response records at any given moment and the question they need to answer in the next moment. Model fit performance is evaluated by comparing the predicted responses to the actual responses. In our datasets, students' responses to questions are recorded sequentially over time. With the learned models, for any given time step, we can estimate the student's knowledge state based on their responses up to that time step, thereby predicting their performance in the next time step. In general, the more accurate the model's prediction of student performance, the more precise its description of the regularities in the cognitive processes.

We test the H-POMDP performance and the POMDP-based cognitive model through the data of {\it ASSIST1}, {\it ASSIST2}, {\it ASSIST3} and {\it Quanlang1}. In the pre-experiment, we find that when the number of cognitive patterns $k$ exceeds $3$, the model's accuracy does not improve further. Hence, we set $k$ to $3$. For each set of knowledge concepts, we conduct ten-fold cross-validation using the following four metrics.
\begin{itemize}
    \item {\bf Prediction Accuracy~(ACC)} is the proportion of samples for which the predicted values match the actual values in the total sample population. In the calculation, the predicted values need to be binarized. The higher the value of ACC, the better the predictive performance.
    
    \item {\bf Area Under an ROC Curve~(AUC)} For a sample set with $M$ positive samples and $N$ negative samples, each positive sample paired with each negative sample forms a sample pair, resulting in $M\times{}N$ sample pairs in total. For these $M\times{}N$ sample pairs, AUC is the proportion of sample pairs where the predicted values of positive samples are greater than the predicted values of negative samples in relation to the total sample pairs. A larger AUC value indicates better predictive performance.
    $$
    \begin{aligned}
        AUC=\frac{\sum{}I(P_{+},P_{-})}{M\times{}N}
    \end{aligned}
    $$where
    $$
    \begin{aligned}
        I(P_{+},P_{-})=
        \begin{cases}
        1,\quad{}\;\;\:{}P_{+}>P_{-} \\
        0.5,\quad{}P_{+}=P_{-} \\
        0,\quad{}\;\;\:P_{+}<P_{-}\\
        \end{cases}
    \end{aligned}
    $$
    \item {\bf Mean Average Error(MAE)} and {\bf Root Mean Squared Error(RMSE)} measure the deviation between the predicted and actual values. A smaller MAE or RMSE value indicates better predictive performance.
    $$
    \begin{aligned}
        MAE&=\frac{\sum{}|y_{actual}-y_{predicted}|}{the\;number\;of\;samples},\\
        RMSE&=\frac{\sqrt{\sum{}(y_{actual}-y_{predicted})^{2}}}{the\;number\;of\;samples}
    \end{aligned}
    $$
\end{itemize}

In Table~\ref{tbl6}, the experimental results show that, for the four sets of knowledge concepts~({\it ASSIST1}, {\it ASSIST2}, {\it ASSIST3} and {\it Quanlang1}), the H-POMDP model outperforms the POMDP-based models across all the four metrics. Specifically, compared to the POMDP-based models, the H-POMDP model exhibits $0.31\%\sim{}1.66\%$ higher in ACC, $0.17\%\sim{}1.88\%$ higher in AUC, $0.05\%\sim{}0.81\%$ lower in RMSE, and $0.71\%\sim{}1.71\%$ lower in MAE.\begin{table}[h]
    \renewcommand{\arraystretch}{1}
    \setlength\tabcolsep{15.625pt}
    \footnotesize
    \caption{Experimental results on the student performance prediction}
    \begin{tabular}{ccrr}
        \toprule
        Sub-dataset & Metric & POMDP & H-POMDP\\
        \midrule
        \multirow{4}{*}{\textit{ASSIST1}} & ACC & 0.6924 & 0.7090\\
        & AUC & 0.7564 & 0.7752\\
        & MAE & 0.3896 & 0.3725\\
        & RMSE & 0.4464 & 0.4386\\
        \\
        \multirow{4}{*}{\textit{ASSIST2}} & ACC & 0.7127 & 0.7157\\
        & AUC & 0.7784 & 0.7850\\
        & MAE & 0.3755 & 0.3684\\
        & RMSE & 0.4365 & 0.4335\\
        \\
        \multirow{4}{*}{\textit{ASSIST3}} & ACC & 0.6975 & 0.7102\\
        & AUC & 0.7609 & 0.7788\\
        & MAE & 0.3823 & 0.3703\\
        & RMSE & 0.4428 & 0.4347\\
        \\
        \multirow{4}{*}{\textit{Quanlang1}} & ACC & 0.7289 & 0.7369\\
        & AUC & 0.7768 & 0.7785\\
        & MAE & 0.3535 & 0.3443\\
        & RMSE & 0.4234 & 0.4229\\
        \bottomrule
    \end{tabular}
    \label{tbl6}
\end{table}

While the H-POMDP models outperform the POMDP-based models in all the metrics for the student performance prediction, their performance on the student population data is only slightly superior. However, these small differences could often be magnified in individual students' learning processes. For instance, a student's struggle with a specific knowledge concept within a set of knowledge concepts can lead to stagnation in their overall learning progress. This situation places greater demands on the model's ability to capture individual students' knowledge states and cognitive patterns and induce effective strategies. We will demonstrate the H-POMDP performance on the learning strategy induction.

\subsection{Experiment 2: Evaluations on Learning Strategy Induction}

We employ IE-DP~\cite{gao2023improving} for the policy induction based on the learned cognitive models. Furthermore, we simulate student learning guided by these learning strategies to evaluate their effectiveness. For each set of knowledge concepts, with the number of interactions set to $10$, we simulate interactions between $10,000$ students and the ITS and record the average performance of the simulated students. Specifically, we record the final knowledge states of the $10,000$ students and calculate the distribution of the states as follows.
$$
\begin{aligned}
    P(s_{T})=sn_{s_{T}}/SN,\quad{}s_{T}\in{}S
\end{aligned}
$$
where $sn_{s_T}$ is the number of students whose final knowledge states are $s_{T}$, and $SN$ is the number~($10,000$) of students in the simulations. We use three metrics to compare the performance of inducing the learning strategies.
\begin{itemize}
\item {\bf Proficiency in the \bm{$k_{th}$} knowledge concept (\bm{$pro(kc_{k})$})} is the average mastery level of the student group in the $k_{th}$ knowledge concept. The higher the value of $pro(kc_{k})$, the better the average mastery level of students in the $k_{th}$ knowledge concept.
$$
\begin{aligned}
    pro(kc_{k})={\sum}_{s_{T}\in{}S}ml(s_{T},kc_{k})P(s_{T})
\end{aligned}
$$
where $kc_{k}$ is the $k_{th}$ knowledge concept and $ml(s_{T},kc_{k})$ is the mastery level~($0$ or $1$) of the student in the $k_{th}$ knowledge concept in the knowledge state $s_{T}$.
\item {\bf Proficiency in all knowledge concepts (\bm{$pro_{sum}$})} is the average number of knowledge concepts mastered by the student group in the learning domain.
$$
\begin{aligned}
    pro_{sum}={\sum}_{s_{T}\in{}S}KC(s_{T})P(s_{T})
\end{aligned}
$$
where $KC(s_T)$ is the number of mastered knowledge concepts in the knowledge state $s_T$.
\item {\bf Variance of the number of knowledge concepts mastered by all students (\bm{$VAR$})} represents the stability of the learning strategy. The lower the value of $VAR$, the more stable the effectiveness of the learning strategy.
$$
\begin{aligned}
    VAR={\sum}_{s_{T}\in{}S}[pro_{sum}-KC(s_{T})]^{2}P(s_{T})
\end{aligned}
$$
\end{itemize}

\begin{table}[h]
    \renewcommand{\arraystretch}{1}
    \setlength\tabcolsep{12.8125pt}
    \tiny
    \caption{Experimental results on the learning strategy performance}
    \begin{tabular}{ccccc}
        \toprule
        Sub-dataset & & & POMDP & H-POMDP\\
        \midrule
        \multirow{9}{*}{\textit{ASSIST1}} & \multirow{3}{*}{\textit{(X,Z,Y)}} & \textit{(0,0,0)} & 0.1011 & 0.0919 \\
         & & ...... & ...... & ......\\
         & & \textit{(1,1,1)} & 0.5652 & 0.8094 \\
         \\
         & \multirow{4}{*}{proficiency} & \textit{X} & 0.7863 & 0.8265 \\
         & & \textit{Z} & 0.5652 & 0.8094 \\
         & & \textit{Y} & 0.7922 & 0.8962 \\
         & & \textit{sum} & 2.1437 & 2.8321 \\
         & & \textit{VAR} & 1.1683 & 0.9874 \\
        \\
        \multirow{13}{*}{\textit{ASSIST2}} & \multirow{3}{*}{\textit{(T,X,U,Y,V,Z,W)}} & \textit{(0,0,0,0,0,0,0)} & 0.0540 & 0.0892 \\
         & & ...... & ....... & ......\\
         & & \textit{(1,1,1,1,1,1,1)} & 0.1123 & 0.2380 \\
        \\
         & \multirow{8}{*}{proficiency} & \textit{T} & 0.6104 & 0.5952 \\
         & & \textit{X} & 0.3139 & 0.3808 \\
         & & \textit{U} & 0.6232 & 0.6554 \\
         & & \textit{Y} & 0.1843 & 0.2563 \\
         & & \textit{V} & 0.7347 & 0.3379 \\
         & & \textit{Z} & 0.1123 & 0.2380 \\
         & & \textit{W} & 0.6142 & 0.7273 \\
         & & \textit{sum} & 3.1930 & 3.1909 \\
         & & \textit{VAR} & 4.2236 & 5.9439 \\
        \\
        \multirow{12}{*}{\textit{ASSIST3}} & \multirow{3}{*}{\textit{(U,X,V,Y,W,Z)}} & \textit{(0,0,0,0,0,0)} & 0.1261 & 0.1021 \\
         & & ...... & ...... & ......\\
         & & \textit{(1,1,1,1,1,1)} & 0.0426 & 0.0359 \\
        \\
         & \multirow{7}{*}{proficiency} & \textit{U} & 0.9316 & 0.5584 \\
         & & \textit{X} & 0.2923 & 0.4455 \\
         & & \textit{V} & 0.6547 & 0.8052 \\
         & & \textit{Y} & 0.3019 & 0.2974 \\
         & & \textit{W} & 0.6444 & 0.4246 \\
         & & \textit{Z} & 0.0426 & 0.0359 \\
         & & \textit{sum} & 2.5675 & 2.5670 \\
         & & \textit{VAR} & 3.2798 & 1.8843 \\
        \\
        \multirow{11}{*}{\textit{Quanlang1}} & \multirow{3}{*}{\textit{(V,W,X,Y,Z)}} & \textit{(0,0,0,0,0)} & 0.1852 & 0.0843 \\
         & & ...... & ...... & ......\\
         & & \textit{(1,1,1,1,1)} & 0.4538 & 0.5865 \\
        \\
         & \multirow{6}{*}{proficiency} & \textit{V} & 0.8148 & 0.9157 \\
         & & \textit{W} & 0.7194 & 0.8964 \\
         & & \textit{X} & 0.6061 & 0.7823 \\
         & & \textit{Y} & 0.6129 & 0.8369 \\
         & & \textit{Z} & 0.4538 & 0.5865 \\
         & & \textit{sum} & 3.2070 & 4.0178 \\
         & & \textit{VAR} &3.9738 & 2.3545 \\
        \bottomrule
    \end{tabular}
    \label{tbl7}
\end{table}\begin{table}[h]
    \renewcommand{\arraystretch}{1.0}
    \setlength\tabcolsep{9.375pt}
    \footnotesize
    \rm
    \caption{The statistical significance of the learning strategy performance}
    \begin{tabular}{lcccc}
        \toprule
        Sub-dataset & {\it ASSIST1} & {\it ASSIST2} & {\it ASSIST3} & {\it Quanlang1} \\
        \midrule
        $t$-value & 26.45 & 0.07 & 0.02 & 32.23 \\
        \bottomrule
    \end{tabular}
    \label{tbl8}
\end{table}

Table~\ref{tbl7} shows the results of the simulated tutoring process. For each dataset~({\it ASSIST1}, {\it ASSIST2}, {\it ASSIST3} and {\it Quanlang1}) and learning strategy, we document the distribution of final knowledge states, e.g. $state(X,Z,Y)$ in {\it ASSIST1}, as seen at the top of the record correspond to each dataset in the table. As an example, for one such record, after students follow the learning strategies induced by the H-POMDP model in the learning domain {\it ASSIST1}, the probability of achieving a final knowledge state of $(1_{X},1_{Z},1_{Y})$ is $0.8094$.

Table~\ref{tbl8} provides the statistical significance of performance difference between the two learning strategies on each dataset. The critical value for the two-tailed $t$-test with a significance level of $0.05$ and the freedom degrees of $19,998$ is $1.96$. Based on the above metrics, we conclude that the learning strategies induced by the H-POMDPs are better.

As shown in Table~\ref{tbl8}, there is significant difference in the performance of these two strategies in {\it ASSIST1} and {\it Quanlang1}, while there are no significant difference in {\it ASSIST2} and {\it ASSIST3}. In {\it ASSIST1} and {\it Quanlang1}, the students who receive learning strategies induced by H-POMDP have a higher average mastery of knowledge concepts compared to the students who receive learning strategies induced by POMDP-based cognitive models. In terms of variance, the former also exhibits larger stability. In {\it ASSIST2} and {\it ASSIST3}, there is little difference in the average mastery of the knowledge concepts under both strategies. In {\it ASSIST2}, the performance of learning strategies induced by POMDP is more stable, whereas in {\it ASSIST3}, H-POMDP performs better. Overall, the performance of learning strategies induced by the H-POMDP models is superior in most cases. It can also be observed that when the knowledge concept structure becomes more complex~(the increasing number of knowledge concepts and the complicated relationships between them), the performance difference between the two becomes less apparent. This is due to the more complex beliefs that the H-POMDP model needs to maintain during the strategy induction, leading to the loss of precision in the approximate solutions.

We report the students' proficiency of mastering each knowledge concept. It is observed that the nodes closer to the root node have a higher probability of being mastered, whereas nodes closer to the leaf nodes have a lower probability of being mastered. For instance, in {\it ASSIST1}, both node $X$ and node $Y$ have a higher probability of being mastered compared to node $Z$. This phenomenon can be attributed to the fact that for a child node to be mastered, its parent node must first be mastered. This is well consistent with the students' learning process.

\section{Related Works}
\label{sec:related}

There has been a long way of examining the feasibility of employing POMDP in the student cognitive modelling and strategy induction within the ITS context. By developing a framework for applying a POMDP model to teaching, Rafferty {\it et al.}~\cite{rafferty2016faster} addressed the issue that deciding what pedagogical decisions to make involves reasoning about a number of different components and balancing conflicting priorities. Clement {\it et al.}~\cite{clement2016comparison} conducted a performance comparison between the POMDP and multi-armed bandits for achieving online planning of effective teaching sequences. They demonstrated the limitations and robustness of each method through simulation.

While the application of POMDPs holds theoretical feasibility in the ITS, the optimization of model parameter learning and strategy resolution becomes imperative when facing complicated scenarios. Wang~\cite{wang2018reinforcement} developed a parameter learning technique that enables a POMDP-based ITS to enhance its teaching capabilities online. Subsequently, the improvement is made to reduce the state space, tree set, and observation set involved in computing the {\it Bellman} equation~\cite{wang2019efficient}. Ramachandran {\it et al.}~\cite{ramachandran2019personalized} designed the Assistive Tutor POMDP~(AT-POMDP) to provide personalized support to students for practicing a difficult math concept over several tutoring sessions. Nioche {\it et al.}~\cite{nioche2021improving} extended the  model-based approaches for ITSs by introducing a modular framework that combines online inference tailored to each user and item with online planning that takes the learner's temporal constraints into account. Gao {\it et al.}~\cite{gao2023improving} developed a POMDP-based cognitive model for the scenario of practice-based learning. They improved cognitive modelling and model parameter learning by incorporating knowledge from knowledge concept structures, and proposed an information entropy-based planning method to induce learning strategies.

In order to facilitate the parameter learning and strategy induction, the existing research has all adhered to the assumption that, for a given learning domain, students possess uniform learning capabilities. However, in reality, most students are different, and they possess different cognitive abilities. In this paper, we propose a novel H-POMDP modelling approach which enables clustering and modelling of students with various learning capacities. Additionally, we develop the techniques for the model parameter learning and strategy induction.

\section{Conclusion and Future Work}
\label{sec:conclusion}

Compared to conventional modelling approaches, personalized modelling is more practical for tutoring individual students on learning knowledge concepts. This paper takes ITS as a research platform and proposes the new model of homomorphic POMDP~(H-POMDP) upon which a novel cognitive modelling approach is developed to induce learning strategies for individual students. The H-POMDP approach allows cognitive models to accommodate multiple distinct cognitive patterns, enabling personalized learning pathways. We encode the practical implications of students' learning patterns so as to improve the model parameter learning, and demonstrate the model performance in multiple problem domains of learning knowledge concepts. 

This work reinforces the strength of POMDP-based approaches to personalize the learning pathway for individual students. The techniques are clearly to represent and explain how the students learn knowledge concepts during the process. We shall note that the current research still focuses on a small set of knowledge concepts in the belief space. It would be very interesting to expand H-POMDPs in more complicated knowledge concept domains. For example, we could integrate the learning of knowledge concepts into the belief space in the H-POMDP model. In addition, this work could be further developed by encoding a set of learning stereotypes in the H-POMDP parameter learning. By doing this, we may discover new cognitive patterns from the learning activity record and improve the H-POMDP reliability on inducing individual learning strategies. 

\bibliographystyle{IEEEtran}
\bibliography{ref}

\end{document}